\documentclass[letterpaper]{article}
\usepackage{flairs}%aaai
\usepackage{times}
\usepackage{helvet}
\usepackage{courier}
\usepackage{graphicx}
\usepackage{tabularx}
\usepackage{booktabs}
\usepackage{multirow}
\usepackage{xcolor}
\usepackage{tabularx}
\usepackage{enumitem}
\usepackage[flushleft]{threeparttable}
\begin{document}
% This file is an adoption of the style file for AAAI Press 
% proceedings, working notes, and technical reports.  This file is made 
% with minimal changes by explicit permission from AAAI.
%
\title{On Information Hiding in Natural Language Systems}
\author{Geetanjali Bihani \and Julia Taylor Rayz\\
Computer and Information Technology\\
Purdue University, USA\\
}
% \author{Author 1 \\}

\maketitle
\begin{abstract}
\begin{quote}
With data privacy becoming more of a necessity than a luxury in today’s digital world, research on more robust models of privacy preservation and information security is on the rise. In this paper, we take a look at Natural Language Steganography (NLS) methods, which perform information hiding in natural language systems, as a means to achieve data security as well as confidentiality. We summarize primary challenges regarding the secrecy and imperceptibility requirements of these systems and propose potential directions of improvement, specifically targeting steganographic text quality. We believe that this study will act as an appropriate framework to build more resilient models of Natural Language Steganography, working towards instilling security within natural language-based neural models.

\end{quote}
\end{abstract}

\section{Introduction}
General purpose neural language models have shown to learn spurious patterns existing within natural language text. Language variety within the training corpora may contain cues that lead to inference-based attacks, increasing the risk of exposure of any private information that has been unintentionally encoded within a given model \cite{nguyen_learning_2021}. The sole reliance on word form and statistical distribution of word tokens within a given training corpora results in lack of comprehension and common-sense reasoning within such models \cite{bender_climbing_2020}. Consequently, the risks that arise from these models can be exploited by adversaries to uncover private attributes of entities mentioned within the text used to train these models, leading to potential fraud and misuse by third parties. To prevent \textit{training data} related privacy leaks, current privacy-preservation techniques focus on training-time updates that utilize adversarial learning, differential privacy-based noise addition or cryptographic enhancements \cite{li_towards_2018,huang_texthide_2020}. Non-cryptographic models preserve privacy through the irreversible removal of rich social signals from the input data \cite{nguyen_learning_2021}. Yet, these models are prone to privacy leakages pertaining to the \textit{training data}, which can be implicit or explicit in nature \cite{huang-etal-2020-texthide}. On the other hand, while current models of cryptographic-enhancements to the training setup can deal with such leakages, their vulnerability to reconstruction attacks deems them inefficient \cite{carlini_is_2021}. 

Apart from distorting or locking the private information through DP-noise addition or cryptographic enhancements respectively, the existence of private information itself can be hidden from the adversaries. Such information hiding techniques have existed parallel to the field of cryptography, within the domain of cybersecurity. Broadly referred to as Steganography, these techniques hide secret information within a given cover medium, and add the component of secrecy on top privacy, leading to broad applications in covert communication, data provenance, etc. \cite{taleby_ahvanooey_modern_2019}. Although steganographic models utilizing natural language have been researched extensively in the past, their application towards \textit{training data} privacy has not been explored.

In this paper, we present a comprehensive review of steganography techniques that use natural language text as a cover medium, i.e. hide data within natural language. These works come under the umbrella of \textit{linguistic steganography (LS)} and \textit{Natural Language Watermarking }\textit{(NLW)} techniques that aim to hide information within the structure and/or meaning of natural language \cite{atallah_natural_2001,abdelnabi_adversarial_2021}. We discuss the limitations of existing approaches and define concrete directions of future research. Our work does not intersect with prior reviews on text steganography approaches because while they focus on methods that alter character level properties of text documents (e.g. character spacing, white spaces, etc.), we specifically focus on steganography methods that employ the modification or generation of natural language. Overall, we make two primary contributions. (1) We summarize approaches towards \textit{NLS} under a unified framework to facilitate future discussions. (2) We critically analyze current approaches and identify potential directions of future research towards privacy and secrecy improvements in \textit{NLS}.

\begin{figure}[htbp]
\centerline{\includegraphics[width=0.45\textwidth]{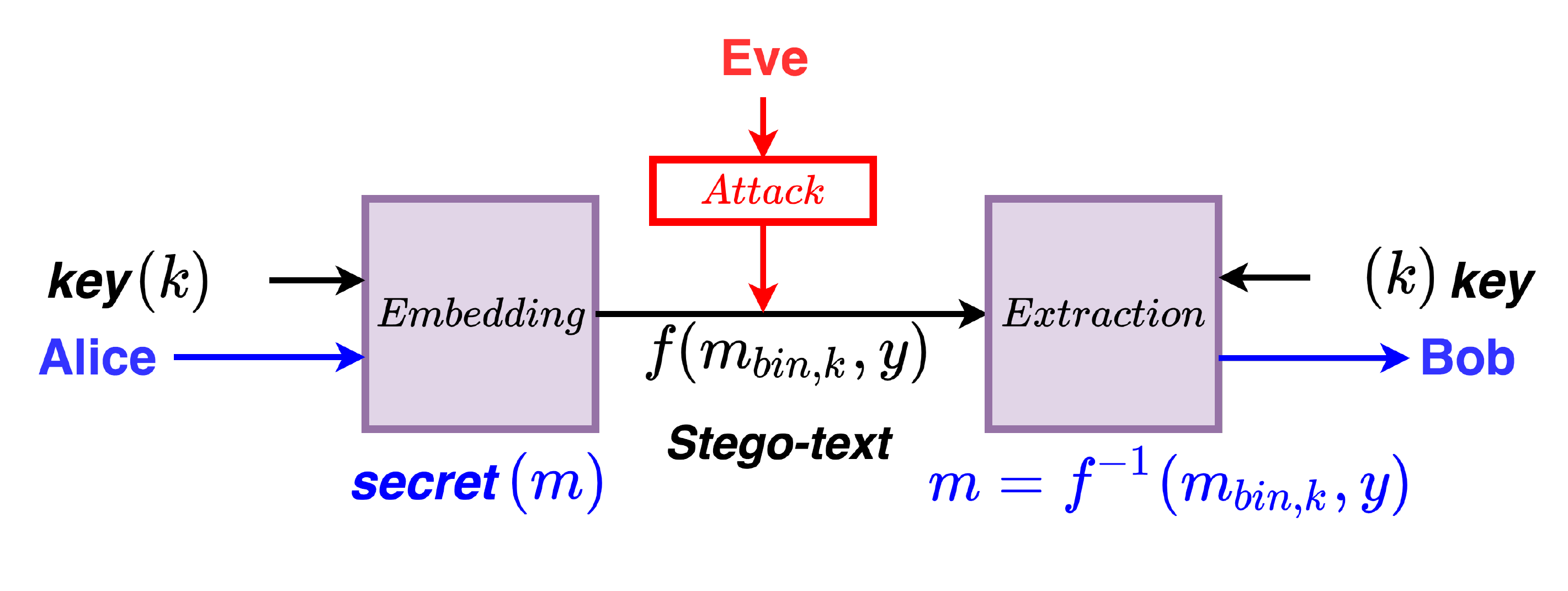}}
\caption{Natural Language Stegosystem}
\label{fig1}
\end{figure}

\section{Natural Language Stegosystem}

In this section, we define various components of a natural language stegosystem, associated attack models, standards for privacy guarantees and NL stegosystem requirements. Formally, the problem statement of steganography is defined through the modified Simmon's Prisoner Problem \cite{simmons1984prisoners}. We consider a steganographic system (\textit{stegosystem}), where two entities \textit{Alice} and \textit{Bob} aim to communicate secret information, without arousing the suspicion of the eavesdropper \textit{Eve}. The overall stegosystem describing the aforementioned scenario is shown in  Figure \ref{fig1}, and described below. 

\subsubsection{Embedding/Encoding $(Emb)$}
This task is employed by \textit{Alice (A)} to encode a secret $(m)$ into a natural language text carrier $(y)$. The secret $(m)$ is first converted into a binary string $(m_{bin})$ and encoded with a secret key $(k)$ to secure the contents of $m$, forming $(m_{bin,k})$. Finally, $(m_{bin,k})$ is embedded within the form, syntax and/or semantics of the carrier $y$, using an invertible function $f$, forming the stego-text $f(m_{bin,k}, y)$.

\subsubsection{Extraction/Decoding $(Ext)$}
This task is employed by \textit{Bob (B)} to decode the secret $(m)$ from the carrier $(y)$. Since Bob has knowledge of the secret key and the encoding function, they can extract the secret $m$ by applying the key $k$ and inverting the function yielding $f^{-1}(m_{bin,k}, y)$ 

\subsubsection{Attack $(Att)$}
This task is employed by \textit{Eve (E)} to break the security of the steganographic protocol using steganalysis techniques. This process comprises of active and passive attacks, where an active attacker aims to destroy the secret while a passive attacker aims to detect the presence of the secret. The attacker is assumed to have no knowledge of the key $k$ used in the embedding procedure as well as limited knowledge regarding the invertible function $f$. 

\subsection{Stegosystem Requirements} 
Written text contains less redundant information as compared to other media such as images \cite{atallah_natural_2001}. This renders the task of hiding information using natural language more challenging because minor changes within the carrier text might result in drastic changes in word meaning, grammatical correctness and language style. Thus, a Natural Language Stegosystem should aim to achieve the following requirements:

\subsubsection{Meaning preservation} This requirement focuses on preserving the meaning of the actual carrier text, while encoding secret information within. Meaning preservation is important in cases where the carrier is also an important component, e.g. NL watermarking.

\subsubsection{Grammaticality} Any stego-text must be grammatically correct in order to attract minimal attention from adversaries. If the adversary can identify gaps in lexical, sentential or text semantics, they might be able to detect the location of the secret within a given text, rendering the stegosystem susceptible to attacks.

\subsubsection{Style Preservation} Language style is an important characteristic of natural language employed by different individuals. Thus, if an NL stegosystem outputs stego-text which doesn't align with the language style employed in the rest of the text, the adversary can easily detect and point towards the location where the stego-secret is embedded, rendering the stegosystem susceptible to attacks.

\subsection{Attack Models}
Different types of adversarial attacks can be conducted on steganographic models of privacy including (i) \textit{White-box attacks:} The attacker has white-box access to the stegosystem, apart from the key, and can misuse these privileges to uncover the secret. (ii) \textit{Black-box attacks:} The attacker has no knowledge of the stegosystem and performs various transformations to break down the stegosystem. (iii) \textit{Grey-box attacks:} The attacker has partial knowledge regarding the stegosystem components and uses this information to strategically attack the stegosystem.

Recent developments in \textit{NLS} utilize general purpose language models \cite{ueoka_frustratingly_2021,ziegler_neural_2019}, which can be easily accessed by an adversary, and increase the risk of grey-box attacks. The robustness of an \textit{NLS} system, against the aforementioned attacks is measured using the imperceptibility metrics. These metrics are divided into two types, i.e.  statistical imperceptibility and human imperceptibility. These imperceptibility metrics measure the risk of detection by human or statistical adversaries respectively.

\section{A Comprehensive Review of NLS}

In this section, we summarize different types of \textit{NLS} techniques, specifically methods that modify or generate natural language text to encode secrets. These techniques can primarily be divided into two types, i.e. Carrier Modification and Carrier Generation techniques. The difference between these methods lies in their treatment of the carrier text used as a medium to encode the secret message $m$. While carrier modification based approaches need to optimize over two goals, i.e. meaning retention and capacity improvement, carrier generation approaches only require optimization over one goal i.e. fine-tuning the carrier generation scheme to generate human-like natural language sequences. Moreover, carrier generation based methods yield higher capacity than carrier modification based approaches. These methods are explained in further detail in the following sections. 
\subsection{Carrier Modification}
Carrier modification techniques modify an existing carrier text written in a given natural language, to encode a secret message. These methods were primarily utilized in many of the initial attempts at \textit{NLS} that hid secrets either through word substitution or modification of the semantic/syntatctic structure of a given cover text.

\subsubsection{Substitution-based Methods} Early substitution based methods for carrier modification used synonyms, chosen to minimize meaning distortion while maximizing ambiguity within the text. The amount of meaning preserved and ambiguity achieved was measured using lexical and sentence level characteristics of the text \cite{topkara_hiding_2006}. The resilience was claimed to be rooted in the adversary's lack of knowledge regarding where the changes were made and the inefficiency of word sense disambiguation techniques during that time.

The use of typographical errors as substitutions for original words has also been explored for hiding information \cite{topkara_information_2007}. The original words are replaced with their typographic error replacements, e.g pa\textbf{r}ty $\rightarrow$ pa\textbf{t}ty. Hiding secrets within abbreviations have also been proposed \cite{shirali-shahreza_text_2007}. Usually, these schemes lack adaptation to language evolution. Since, old abbreviations might not be used as language evolves over time, these schemes end up being redundant and inefficient.

Recent developments in language models have eliminated the need for rule construction and allow quick access to alternative word choices for substitutions. General purpose language models have been utilized to perform word substitution using various masking-based encoding schemes \cite{ueoka_frustratingly_2021} as well as adversarial training \cite{abdelnabi_adversarial_2021}. By training an adversarial model that performs classification between input and modified text, the model aims to bring stego-text as close to natural language text, while embedding the secret. 

\subsubsection{Language Structure-based Methods}Early research towards carrier modification based \textit{NLS} also utilized the structure of natural language itself to hide information. These methods encode the secret message within the syntactic choices or text meaning representations of a given carrier text \cite{atallah_natural_2001}. Proposed as an alternative to the primarily researched substitution techniques, these methods showed immunity against substitution-based attacks. Concurrent methods of semantic transformations to encode secret message/watermark in texts have also been researched \cite{atallah_natural_2003}. Thus, even if the adversary knows the steganography scheme itself, they would not be able to identify the specific modifications in sentence structure where the secrets have been encoded, wherein lies the resilience of these methods.

\subsection{Carrier Generation}
Carrier Generation techniques aim to generate natural language carrier texts that encode secret messages. These techniques range from carrier generation using Context-Free-Grammars (CFGs) \cite{wayner1992mimic,chapman1997hiding}, Markov chains \cite{shniperov_text_2016} and neural networks \cite{ziegler_neural_2019,yang_vae-stega_2021}. The increased computational prowess and improved language modeling capabilities have resulted in major improvements in carrier generation models of \textit{NLS}. Here, we divide carrier generation based \textit{NLS} approaches into three categories, i.e. \textit{(CFG)} based methods, Conditional Probability \textit{(ConProc) }based methods and latent space semantics based methods.

\subsubsection{Context-Free Grammars}
One of the earliest attempts towards generating natural language texts for \textit{NLS} proposed the use of context-free-grammars (CFGs) \cite{wayner1992mimic}. A CFG is a set of production rules, forming the formal grammar for a given language, which can universally describe any combination of valid text sequences in that language. Thus, assuming that the CFG encompasses all possible text sequence permutations present within a given natural language, it can be used to generate syntactically legitimate text sequences. In order to hide information within the text generated using CFGs, Wayner developed a custom-made CFG where the choice of the CFG branch portrayed the bit(s) encoded \cite{wayner1992mimic}. While these grammars can yield syntactically correct outputs, they can lead to repeated sentences, unless huge grammars are designed \cite{chapman1997hiding}. Towards generating semantically correct outputs, \cite{chapman1997hiding} combined a dictionary table containing a large list of POS tags, word synset and word pairs, and style templates to generate cover texts. But these methods have mostly been overtaken by conditional probability based frameworks.

\subsubsection{Conditional Probability based Frameworks \textit{(ConProc)}} Statistical models of language usually consider sentences as a sequence of words or a sequence signal. This can be modeled by considering the conditional distribution probability of each word in the text corpora \cite{bengio2003neural}, shown in the following equation:

$$ {P}\left(S\right)=\prod_{t=1}^{n} {P}\left(w_{t} \mid w_{1}^{t-1}\right)$$

$$ {P}\left(w_{t} \mid {context}\right) \forall {t} \in {V} $$

Here,  $S$ denotes the sentence of length $n$, $w_{t}$ is the t-th word in the sentence and $w_{i}^{j}=\left(w_{i}, w_{i+1}, \cdots, w_{j-1}, w_{j}\right)$. Once this distribution is learnt by a statistical language model for all the training instances, the model can be used to generate word sequences based on the previous words encountered. Prior works have utilized this conditional probability \textit{(ConProc)} framework to generate carrier text that encodes secret messages within the conditional distribution of words in language. These methods include works done using Markov models as well as neural networks.

\begin{itemize}
    \item \textit{Markov Models: } Initial attempts towards carrier generation using statistical language models involved the use of markov chain models \cite{shniperov_text_2016}. Markov chain models are stochastic models that describes a sequence of possible events, where the probability of each event solely depends on the previous event state. For carrier generation, the Markov chain model is used to calculate the transition probability, based on the number of occurrences of different text sequences. This transition probability is then used to encode words and therefore secret bits during the process of carrier generation. 

    \item \textit{Neural Language Models (ConProc): } The initial neural steganographic carrier generation models primarily employed LSTMs, due to their state of the art performance in text generation \cite{yang_rits_2018,yang_rnn-stega_2019}. Recent developments in neural language models have led to substantial improvements in neural text generation. In a recent work \cite{ziegler_neural_2019}, the authors propose to utilize language models and arithmetic coding to generate cover text for text steganography. Comparison between  variable length coding and fixed length coding also revealed that variable length coding produces generates better carriers in terms of statistical imperceptibility \cite{dai-cai-2019-towards}.

Since conditional distribution of word tokens over text is not necessarily uniform at all times, models that randomly choose next tokens can lead to generating rare text sequences, which can distort the original text distribution, and increase the risk of adversarial attacks. To address this shortcoming, \cite{dai-cai-2019-towards} proposed the \textit{`patient-huffman'} algorithm, which waits for an appropriate opportunity to encode secret bits in text, where ambiguity is higher and candidate tokens are more uniformly distributed. 

In order to improve the human imperceptibility in generated stego-texts, \cite{yang_linguistic_2021} propose a method to constrain the semantic expression of a generated text by balancing between human imperceptibility and statistical imperceptibility. This was achieved by utilizing an encoder to learn `semantic' information of the context and a decoder to generate natural text corresponding to the specific `semantic' information. In a related stream of works, some authors have also utilized Chinese poetry to perform carrier generation based text steganography \cite{luo_text_2016}. These works use a template-constrained generation method and utilize inner-word mutual information to choose words to create text sequences, encoding the secret message within quatrains, a specific genre of Chinese poetry. Although an interesting avenue of research, the authors add that the generated text lacks logical relations between sentences.

\end{itemize}

\subsubsection{Latent Space Steganography }Apart from utilizing the \textit{ConProc} framework to generate steganographic text, a recent work proposed utilizing a latent semantic space to encode secrets \cite{zhang_linguistic_2020}. The model maps the secret message to a discrete semantic space, defined by natural language semantemes (themes/topics), and the corresponding semantic vector $\alpha$ is fed to a conditional text generation model, where the model generates stegotext $x$ conditioned on $\alpha$. The generated stego-text is relevant to the semanteme and sent to the receiver, who can use a semantic classifier to decode the stegotext. This work is one of the first approaches towards latent space steganography, where the secret message is encoded with a latent space and mapped to the symbolic space. The authors propose that hiding secrets in an implicit manner can lead to better concealment, as long as the prior distribution of the latent space remains unchanged \cite{zhang_linguistic_2020}.

\section{Challenges in NL Steganography} \label{sec2}
Natural language steganography has evolved over time, from CFG-based text constructions \cite{wayner1992mimic,chapman1997hiding} to neural language models for text generation and modification \cite{yang_rnn-stega_2019,yang_vae-stega_2021,ziegler_neural_2019,ueoka_frustratingly_2021}. Although these developments have pushed the field further in terms of steganographic quality improvements, many gaps remain to be addressed. In this section, we describe some of the primary challenges faced by the field of natural language steganography in the recent years.

\subsection{Lack of anti-steganalysis capabilities}

Over the last few decades, linguistic steganalysis approaches have been developed alongside NL steganography models. These approaches aim to differentiate between steganographic and non-steganographic carrier texts. Recent methods extract and compare difference in word dependencies across steganographic and natural texts using neural networks \cite{bao2020text,yang2019fast}. Current NL steganography works do not account for resilience against the aforementioned steganalysis approaches. These works ignore the prerequisite of defining the anti-steganalysis abilities of their proposed models, and reduce model evaluation to statistical or human imperceptibility metrics such as KL-divergence and BLEU. Moreover, the works that do perform anti-steganalysis tests \cite{yang_vae-stega_2021,zhang_provably_2021}, exhibit a higher than chance probability of the steganographic text being detected by anti-steganalysis methods. 

\subsection{Data dependent models lack grounding}
Recent developments in neural architectures, easy access to large text corpora and improved computational capabilities have led to the rise of neural language modeling techniques. Although these language models have portrayed state-of-the-art performance on various language generation tasks, their sole reliance on word form to learn the properties of language has been criticized \cite{bender_climbing_2020}. Unlike knowledge-based approaches, where entity relations are explicitly defined, neural models fail to recognize knowledge relations present within the given text, be it conceptual categories or linguistic relations \cite{bihani_low_2021}. These limitations have already been delineated in several recent works that explore the definition of the `meaning' learnt by neural and statistical language models \cite{bender_climbing_2020}. Yet, current carrier modification and generation based NL steganography works do not account for these developments in linguistic semantics and language grounding. As a result, neural network based carrier generation has shown to produce inconsistent text in terms of factual accuracy \cite{ziegler_neural_2019}, increasing the risk of adversarial attacks.

\subsection{Lack of text coherence}
NL steganography approaches perform optimization of the carrier text on the basis of various imperceptibility features, including semantic coherence. If a steganographic carrier text lacks the general semantic characteristics followed by the neighbouring non-steganographic texts, it leads to a higher chance of detection by statistical as well as human adversaries. Unfortunately, current NL steganography approaches limit themselves to the optimization of sentence-level semantic coherence, and do not account for document level semantic coherence \cite{ziegler_neural_2019,yang_vae-stega_2021}. This results in production of text where each sentence seems valid in isolation, but lacks fluency when viewed as a part of a document. 

\subsection{Lack of standardization}
Although there exists a large body of research on NL stenographic techniques, these methods do not establish benchmarks regarding model evaluation, with works varying drastically in terms of datasets used to train and validate the models and metrics used to perform evaluation. Creating steganographic text that is imperceptible to adversaries is the key goal in the field of NL steganography. Given that the adversary can be human or statistical, the steganographic protocols need to produce outputs that are imperceptible to both. Solely focusing on the improvement of one imperceptibility criterion is not enough. Recent research points towards the existence of PSIC effect \cite{yang_vae-stega_2021}, portraying an inverse relationship between human and statistical imperceptibility. Given these priors and towards the goal of overall imperceptibility improvements in NL steganography, we enlist gaps in the evaluation of steganographic text imperceptibility in NL steganography.

\subsubsection{Human Imperceptibility Evaluation}
Human imperceptibility evaluation metrics are highly scattered across works in NL steganography. These metrics include several MT evaluation metrics such as ROUGE and METEOR \cite{yang_vae-stega_2021,yang_linguistic_2021}, etc. The use of MT evaluation metrics for human imperceptibility evaluation cannot be justified because such metrics prefer that the transformed text sequences match the length of the original text sequence, which is not a rigid requirement for NL steganography \cite{chang_secrets_2012}. Moreover, there exist no studies that confirm that MT evaluation metrics are highly correlated with high human imperceptibility judgements on steganographic texts. Additionally, there is a lack of consensus regarding the details of the human annotation tasks, where some papers ask the annotators to evaluate whether the stego-text was written by an actual human being \cite{luo_text_2016}, while others require the annotators to judge whether a generated text is contextually relevant \cite{ziegler_neural_2019}. These differences make it difficult to compare existing approaches, and demand more research on developing standardized metrics that should be used to optimize carrier generation. Thus, there needs to be better standardization of the human imperceptibility evaluation tasks, including the development of a standard definition of human imperceptibility, theoretically and empirically grounded metrics and methods for measuring it, and studies regarding whether there exists a correlation between MT evaluation metrics and human imperceptibility of steganographic texts.

\subsubsection{Statistical Imperceptibility Evaluation}
Unlike human imperceptibility evaluation, statistical imperceptibility evaluation has undergone a more consistent development, in terms of the metrics and methods used to measure imperceptibility. While some recent papers entirely ignore to report any statistical imperceptibility results \cite{yang_linguistic_2021}, others have largely been limited themselves to KL divergence and classification based attacks \cite{ziegler_neural_2019,yang_vae-stega_2021}. Current literature deals with each metric in isolation, and to our knowledge, works performing evaluations on both the criteria do not exist.

\section*{Conclusion}
In this work, we systematize existing \textit{Natural Language Steganography} approaches and outline several challenges. These challenges pertain to the lack of benchmarking, minimal iterative testing and lack of steganographic carrier quality as compared to human-generated texts in current \textit{NLS} approaches. Unlike steganography in other media, \textit{NLS} needs to account for the additional requirement of artificially creating legitimate natural language sequences, a task that yet to be completely solved. We hope that this review on utilizing language as a medium to hide private information can facilitate future research on privacy preservation in language models.

% \section{ Acknowledgments}
% This work is supported by the Ross-Lynn Graduate Student Fellowship.

\bibliography{references.bib}
\bibliographystyle{flairs}
\end{document}